\PassOptionsToPackage{backgroundcolor=white}{todonotes}
\documentclass[letterpaper, 10 pt]{ieeeconf} 
\usepackage[final]{changes}
\usepackage{bm}
\usepackage{amsmath}
\usepackage{amsmath,amssymb}
\usepackage{xspace}
\usepackage[percent]{overpic}
\usepackage{booktabs,floatrow}
\usepackage[font=small]{caption}
\usepackage{wrapfig}
\usepackage{bbm}
\usepackage{dsfont}
\usepackage[capitalize]{cleveref}
\usepackage{caption}
\usepackage{subcaption}

\IEEEoverridecommandlockouts                              

\DeclareMathOperator*{\argmin}{arg\,min}

\title{\LARGE \bf
Learning to Manipulate Tools by \\ Aligning Simulation to Video Demonstration
}

\author{Kateryna Zorina, Justin Carpentier, Josef Sivic and Vladim\'ir Petr\'ik
\thanks{This work was funded by the European Regional Development Fund under the project IMPACT (reg. no. CZ.02.1.01/0.0/0.0/15\_003/0000468), the Grant Agency of the Czech Technical University in Prague, grant No.  SGS21/178/OHK3/3T/17, the French government under management of Agence Nationale de la Recherche as part of the "Investissements d'avenir" program, reference ANR-19-P3IA-0001 (PRAIRIE 3IA Institute) and the Louis Vuitton ENS Chair on Artificial Intelligence.
}
\thanks{
K. Zorina, J. Sivic and V. Petr\'ik are with Czech Institute of Informatics, Robotics and Cybernetics, Czech Technical University in Prague {\texttt{\{kateryna.zorina, josef.sivic, vladimir.petrik\}@cvut.cz}}.}%
\thanks{J. Carpentier is with 
Inria Paris and Departement d'informatique de l'ENS, \'Ecole normale sup\'erieure, CNRS, PSL Research University, 75005 Paris, France {\tt justin.carpentier@inria.fr}.}
}

\makeatletter
\DeclareRobustCommand\onedot{\futurelet\@let@token\@onedot}
\def\@onedot{\ifx\@let@token.\else.\null\fi\xspace}

\def\eg{\emph{e.g}\onedot} 
\def\ie{\emph{i.e}\onedot}

\makeatother

\begin{document}

\maketitle
\thispagestyle{empty}
\pagestyle{empty}


\begin{abstract}
A seamless integration of robots into human environments requires robots to learn how to use existing human tools.
Current approaches for learning tool manipulation skills mostly rely on expert demonstrations provided in the target robot environment, for example, by manually guiding the robot manipulator or by teleoperation.
In this work, we introduce an automated approach that replaces an expert demonstration with a Youtube video for learning a tool manipulation strategy.
The main contributions are twofold.
First, we design an alignment procedure that aligns the simulated environment with the real-world scene observed in the video. This is formulated as an optimization problem that finds a spatial alignment of the tool trajectory to maximize the sparse goal reward given by the environment.  
Second, we describe an imitation learning approach that focuses on the trajectory of the {\em tool} rather than the motion of the human. For this we combine reinforcement learning with an optimization procedure to find a control policy and the placement of the robot based on the tool motion in the aligned environment.
We demonstrate the proposed approach on spade, scythe and hammer tools in simulation, and show the effectiveness of the trained policy for the spade on a real Franka Emika Panda robot demonstration. 
\end{abstract}


\section{Introduction}
\begin{figure}[t]
  \centering
  \includegraphics[width=1\linewidth]{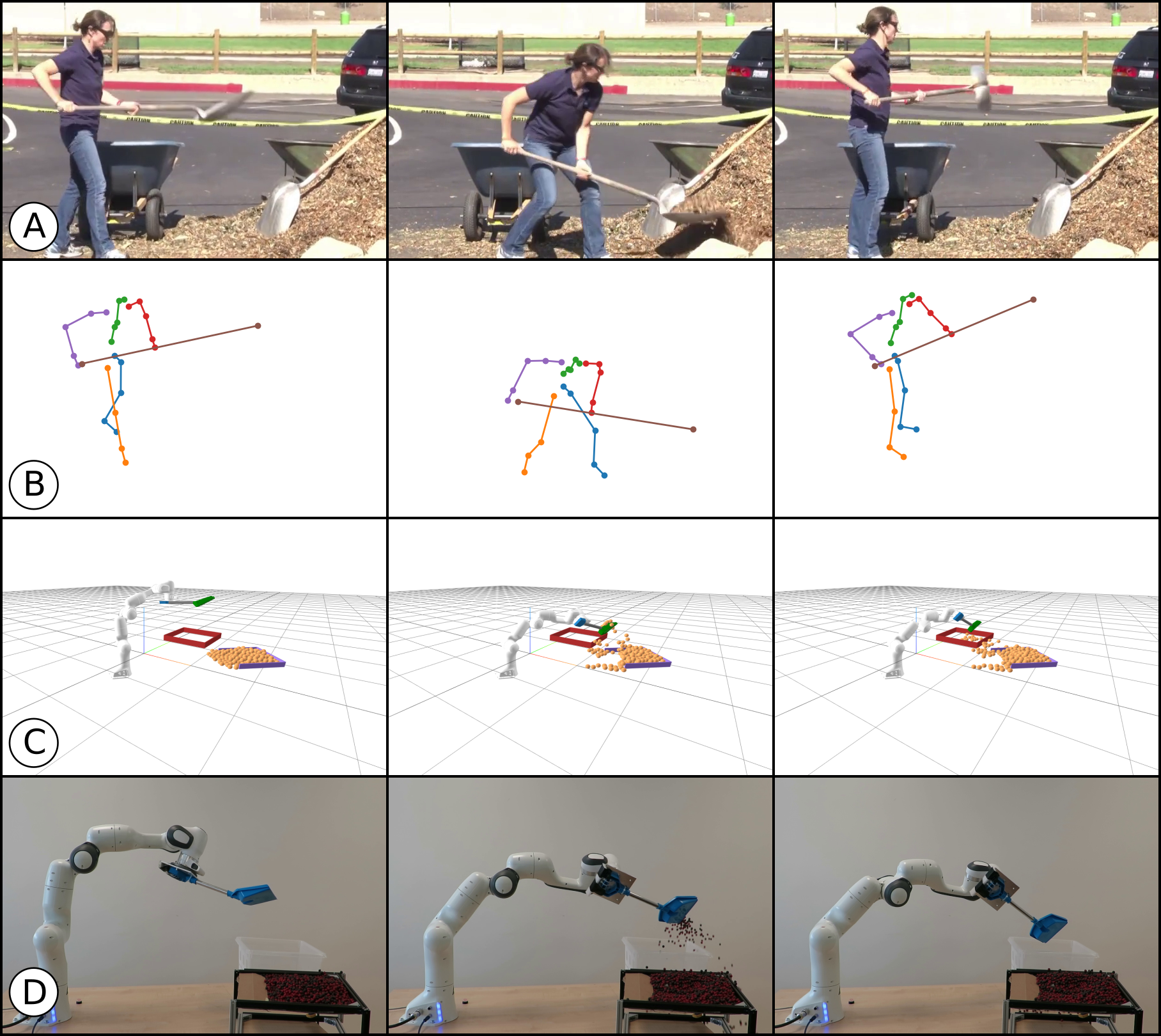}
  \caption{
  {\bf Learning tool manipulation from unconstrained instructional videos} here shown on 
  learning the spade task policy for the Panda robot.
  \vspace{-5mm}
  }
  \label{fig:teaser}
\end{figure}

Robotic systems are an essential part of the modern world.  Robots assemble products on manufacturing lines, transport goods in warehouses, or clean floors in our living rooms. However, outside of controlled settings, robots are far behind humans in terms of dexterity and agility when it comes to using human tools in uncontrolled environments~\cite{4141029}.
Existing approaches to designing robotic skills in human environments, for example, pouring water~\cite{4115573, caccavale_kinesthetic_2019}, using kitchen tools~\cite{5723326}, drilling~\cite{atkinson2007robotic}, or hammering~\cite{fitzgerald_human-guided_2019}, rely on complex motion planning~\cite{4115573, 5723326, atkinson2007robotic} or on learning from expert demonstrations provided in the robot's environment by manually guiding the robotic manipulator~\cite{caccavale_kinesthetic_2019, fitzgerald_human-guided_2019}.
Manual demonstration or teleoperation are costly and robot-dependent, limiting the scalability of the approach, especially when transferring skills to different robotic platforms such as industrial manipulators or humanoids.
This work aims to replace the manual demonstration by information extracted from an instructional video that can be found on the Internet (Fig.~\ref{fig:teaser},~A).
Leveraging information from online instructional videos opens up the exciting possibility of quickly learning a wide variety of new skills without the need for costly manual demonstrations or expert motion programming.

\begin{figure*}[t]
  \centering
  \vspace*{5pt}
    \includegraphics[width=\linewidth]{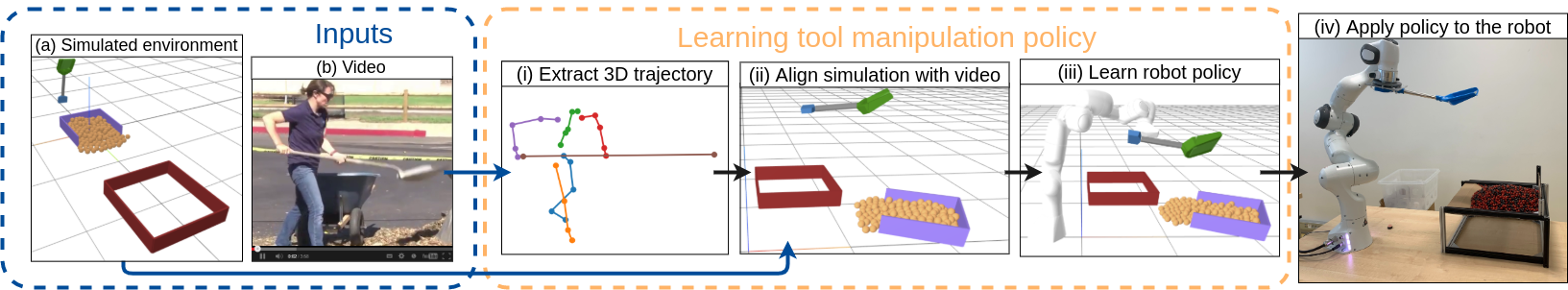}
  \caption{
  {\bf An overview of the proposed approach.} Input consists of (a)~a simulated environment and (b)~an input video demonstration.
  The proposed approach proceeds along the following four steps:
  (i)~extract the 3D trajectory of the human and the tool from the input video using~\cite{DBLP:journals/corr/abs-1904-02683}, 
  (ii)~align the simulated environment with the input video using the extracted tool trajectory (Sec.~\ref{subsec:trajectory_alignment}), 
  (iii)~learn robot policy in the simulated environment using reinforcement learning guided by trajectory optimization (Sec.~\ref{subsec:policy_learning}),
  and (iv)~execute the learned policy on the real robot. The two main technical contributions of this work are in steps (ii) and (iii).
  \vspace{-5mm}}
  \label{fig:approach}
\end{figure*}

For extracting information from videos, we use a motion reconstruction approach~\cite{DBLP:journals/corr/abs-1904-02683} that provides an automatic reconstruction of the whole human body and the tool motion demonstrated in a video. 
In this work, we describe a generic approach for transferring this extracted motion to a robotic system to solve a tool manipulation task by a robot.
We assume that we have access to a simulated environment for the task with a sparse reward function that indicates the completion of the task. 
Such environments can be constructed using standard simulation tools such as~\cite{physx}. 
Using only the sparse reward signal for learning the tool manipulation policy is extremely difficult, as we also show in Section~\ref{subsec:evaluation}. 
We demonstrate that a good policy can be learned by using an Internet video as a demonstration to guide the learning process. 
Using the human video demonstration for learning the tool manipulation task presents the following two technical challenges: 
First, how to adjust a simulated environment to approximately resemble the scene in the video? 
Second, how to map the human motion to the robot morphology, which is also known as motion retargeting? 
We present the following contributions that address these challenges. 
First, we design an alignment procedure that aligns the simulated environment with the real-world scene observed in the video. This alignment procedure uses the trajectory extracted from the video to condition sampling of the scene object positions and properties as well as the unknown tool rotation.
Second, we combine reinforcement learning with an optimization procedure to find a control policy and the placement of the robot based on the tool motion in the aligned environment.
The overview of the proposed approach is shown in Fig.~\ref{fig:approach}.
Focusing on the tool trajectory has the advantage of being independent of the kinematic structure of the robot and allows us to easily train control policies for different robot morphologies.
We illustrate the effectiveness and versatility of our approach on three different robots: Franka Emika Panda, UR5, and the Talos humanoid robot with a fixed lower body, applied for three different manipulation tasks involving various tools: spade, hammer, and scythe. Additionally, we show the transferability of the trained spade policy to the real Franka Emika Panda robotic arm~\cite{franka-panda-robot}, as shown in Fig.~\ref{fig:teaser}.
lease refer to our project page~\cite{project_page} for the supplementary video and code.

\section{Related work}
\label{sec:related_works}

\noindent \textbf{Reinforcement learning}~(RL) methods have been applied in robotics for solving various robotics tasks, for example,  helicopter control~\cite{abbeel_application_2007}, putting a ball in a cup~\cite{NIPS2008_3545}, pouring water into a glass~\cite{4115573}, stacking blocks~\cite{riedmiller_learning_nodate} or solving Rubik's cube with a robotic hand~\cite{openai2019solving}. Solving these tasks usually requires having a dense reward that is obtained by manual reward shaping. 
Even with a dense reward, the policy search often falls in a local optima.
Such poor local optima can be avoided by using a guided policy search~\cite{levine2013GPS} that guides RL using samples from a guiding distribution constructed with trajectory optimization methods. We build on this line of work and consider powerful trajectory optimization techniques as a way to initialize reinforcement learning.
Guided policy search was successfully applied, for example, for end-to-end training of policies that map input images to robot motor torques~\cite{levine2016GPSvisuomotor} for tasks such as  hanging a coat hanger on a clothes rack or fitting the claw of a toy hammer under a nail.  
However, many problems, including the ones tackled in our work, are specified only by a sparse reward, in which the rewarding signal is given only after the task is successfully completed. Solving tasks with sparse rewards remains a challenging open problem because of the need for complicated environment exploration. This problem can be addressed, for example, by using smaller auxiliary tasks along with a sparse goal reward and a scheduler that combines smaller tasks into a sequence of actions~\cite{riedmiller_learning_nodate}. However, this approach relies on computing the auxiliary rewards, which may be environment dependent and requires substantial domain knowledge. 
Another way to improve exploration in environments with sparse rewards is to learn from demonstrations, which we also explore in this work.

\noindent \textbf{Learning from demonstration} can initialize the agent and speed up the optimal policy search~\cite{ARGALL2009469}. It has been successfully shown on various tasks~\cite{fitzgerald_human-guided_2019, NIPS2008_3545, lfd1, lfd2,  lfd_in_rl}.  Examples include learning Dynamic Movement Primitives (DMPs)~\cite{fitzgerald_human-guided_2019}, that can be further adjusted to unfamiliar tools using human demonstrated corrections. Others have used a combination of DMPs and RL to learn in-contact skills~\cite{lfd1} or have included demonstrations into the replay buffer along with the agent's experience~\cite{lfd_in_rl}. 
In another example~\cite{abbeel2010autonomous}, a probabilistic approach  is used to find an intended trajectory from multiple expert demonstrations to guide the design of a controller for autonomous helicopter flight.
For all these methods, however, the demonstrations are given in the agent's environment.  In contrast, we aim at replacing the demonstration in the agent's environment by a demonstration in an instructional video downloaded from the Internet. 

\noindent \textbf{Learning skills from videos.}
Using videos to improve the efficiency of RL has been studied in the past for human or animal motion skills ~\cite{DBLP:journals/corr/abs-1804-02717, RoboImitationPeng20, 2018-TOG-SFV}. DeepMimic~\cite{DBLP:journals/corr/abs-1804-02717} uses a reference motion in combination with the task-specific goal to learn a set of motion skills. Reference motions are presented as a sequence of character target poses obtained by manual pre-processing of mocap clips.
SFV~\cite{2018-TOG-SFV} uses human poses automatically estimated from the input video. The agent learns to replicate the reference human motion.
Related is also~\cite{RoboImitationPeng20} that learns from animal videos and re-targets the observed motion to the simulation. 
Instead of extracting from the video and re-targeting the motion of the human, we focus on the motion of the tool. 
This allows transferring the learned skills to robots of varying morphologies that differ from the morphology of the human observed in the video. 
Another approach~\cite{ImitationFromObservation} uses “imitation-from-observation” that translates the available expert demonstration to the current robot context. This allows to compute a reward signal in an unseen environment and apply RL to learn a control policy.
In contrast, we use video demonstration for policy initialization rather than computing reward based on demonstration.
Another direction for learning skills from videos is using representations learned from unlabeled videos to compute reward functions for RL~\cite{sermanet2018time, mees2020adversarial}.
Learning the representation typically requires recorded or generated training sets (133 sequences for a pouring task in~\cite{sermanet2018time}, 300 simulated videos, and 60 real-world videos per skill in~\cite{mees2020adversarial}). In contrast, we only require a single input demonstration video. This is possible because we extract the 3D motion of the tool and the person from the video using the approach developed in~\cite{DBLP:journals/corr/abs-1904-02683} that uses pre-learnt generic person and object detectors.

\noindent \textbf{Estimating 3D pose from video.} In this work, we use~\cite{DBLP:journals/corr/abs-1904-02683} for estimating the 3D motion of the human and the tool in the video. While other methods often focus on reconstructing only the 3D pose of the object, typically from a single input RGB image~\cite{ brachmann_uncertainty-driven_2016, grabner_3d_2018,rad_bb8_2017, xiang_posecnn_2018}, the approach described in~\cite{DBLP:journals/corr/abs-1904-02683} provides a 3D trajectory of the tool (spade, hammer) manipulated by a human, where the tool trajectory is estimated jointly with the trajectory of the human. 

\noindent \textbf{Robot-tool interaction} research has addressed a range of problems including identifying suitable tools for a certain task~\cite{brown_tool_2011, tee_towards_2018}, tool grasping~\cite{hoffmann_adaptive_2014,wang_rf-compass_2013} or tool guidance using visual feedback~\cite{kemp_robot_nodate}.
In contrast, we focus on learning to imitate the full 3D trajectory of the tool extracted from an input video and assume that the tool is rigidly connected to the target robot manipulator. Given the focus on the tool, the morphology of the target robot can be easily changed.

\section{Learning Tool Manipulation Policy}
\label{sec:approach}

Our aim is to learn a policy for the robot that manipulates a tool to complete a specified task, as illustrated in Fig.~\ref{fig:approach}.
The input of the proposed approach consists of: (a)~a simulated environment with a sparse reward signal $r^g$ and (b)~an instructional video of a human operating the tool in a real-world scene.
This approach is generic and can be used in various environments, but in the following we will use the spade environment as a running example to simplify the explanation. More environments are shown in Sec.~\ref{sec:experiments}.

The simulated environment is parameterized by a set of parameters $\mathcal{P}$ that include the positions and other properties (\eg rotation) of the scene elements. The exact values for parameters $\mathcal{P}$ are not known \textit{a priori}, and only upper and lower limits for these parameters are provided.

The proposed approach (see Fig.~\ref{fig:approach}) starts from the instructional video. 
We (i)~extract the tool motion from the video and apply an (ii)~alignment step to transform the simulated environment to approximately resemble the setup shown in the input video. This allows us to use the extracted trajectory as a guiding signal via trajectory optimization in the subsequent (iii) reinforcement learning, which finds a control policy for the robot.     
Finally, we (iv)~transfer the sequence of actions from the simulated environment to the real robot. These steps are described in detail next. 

\subsection{Extracting the tool trajectory from the video}
Our input is an instructional video of the tool manipulation performed by a human. 
We use the approach described in~\cite{DBLP:journals/corr/abs-1904-02683} to extract the 3D trajectories of the human body and the manipulated tool from the video. The approach combines visual recognition techniques with trajectory optimization to find the 3D trajectory of the tool and the human that best explains the observed input video while modeling contact interactions and full body dynamics.      
The tool is represented as a line segment and the tool trajectory is represented by the 3D positions of the line segment's endpoints over time.
Modeling the tool as a simple 3D line segment is general and encompasses a large set of tools manipulated by humans, including hammers, shovels, or rakes. Given this simplified model, the tool's orientation is not completely determined because the rotation about one axis (along the length of the tool) is unknown. This unknown parameter can be determined, together with other unknown parameters of the environment in the subsequent alignment phase.  

\subsection{Aligning the simulated environment with the input video}
\label{subsec:trajectory_alignment}
From the video, we extract the 3D trajectory of the  tool, and the sequence of 3D human poses. 
However, we do not extract any information about the environment and how the tool is related to the other scene elements.
For the spade example shown in Fig.~\ref{fig:approach}, the position of the sand pile or the target location where the sand needs to be transferred are unknown. 
Hence, aligning the simulated environment with the input video scene constitutes a major challenge. 
To deal with this challenge, we have designed an alignment procedure that proceeds along the following  steps: (i) we construct distribution $\mathbb{P}$ for parameters $\mathcal{P}$ of the simulation (\eg~the placement of the sand and the target box) based on the trajectory extracted from the video; (ii) we sample the simulation parameters from the constructed distribution, (iii) we follow the extracted trajectory in simulated environment and observe if there is any goal reward (\eg~has any sand been transferred to the target box). A positive goal reward means that our simulation with the currently sampled parameters $\mathcal{P}$ is a sufficient approximation of the scene from the video. 
In detail, the goal is to find a set of parameters $\mathcal{P}$ that leads to non-zero goal reward $r^g$, where $r^g$ is obtained from the environment after following the extracted trajectory.

The main idea of the proposed alignment approach is that a suitable distribution of parameters of the simulated environment can be constructed by analyzing the trajectory of the tool because the tool interacts with the important elements of the environment. Hence, we extract keypoints along the tool trajectory and place the task-related scene elements (\eg~the sand deposit or the goal box) at candidate locations given by those keypoints.

\vspace*{1mm}
\noindent \textbf{Identifying keypoints along the trajectory.} 
We assume that the trajectory has keypoints at positions in which the tool interacts with the important scene elements in the environment. These keypoints will act as candidates for the placement of the scene elements. 
We identify a set of keypoints $\mathcal K = \{\bm{k_1},\dots,\bm{k_m}\}$ as points at which the tool velocity is high (top 5\%) or low (bottom 5\%).

\vspace*{1mm}
\noindent  \textbf{Sampling scene elements.} The positions of the scene elements $\bm x_j$ are sampled randomly from a Gaussian mixture with means located at the computed keypoints~$\bm{k_i}$ and tunable covariance matrix~$\Sigma$: $ \bm x_j \sim \sum_{i=1}^m w_i \mathcal{N}(\bm{k_i},\Sigma)$, where $w_i$ is a weight of the corresponding component. To avoid sampling two objects too close to each other, we define a minimum distance between objects as: $d_{th} = 0.1 (d_{\text{max}}-d_{\text{min}})$ where $d_{\text{max}}$ is the maximum and $d_{\text{min}}$ is the minimum distance between any two points of the entire extracted tool trajectory.

\vspace*{1mm}
\noindent  \textbf{Sampling unknown tool rotation, tool offset and other parameters.} We sample the unknown tool rotation \mbox{$\alpha \sim \mathcal{U}(-\pi;\pi)$}. We assume that the rotation of the tool can only change at keypoints, and  is constant on the segments between two keypoints.
The position offset of the tool along the $z$ axis is sampled as $p_z \sim \mathcal{N}(0,0.5)$. 
In addition, we sample a trajectory scale uniformly from the values $\{0.5, 0.75, 1\}$.
The scale is used to resize the trajectory to allow a smaller robot to reach each of the trajectory points without violating kinematic constraints (\eg~joint limits).
Other parameters, denoted~$p_i$, are sampled from the uniform distribution $p_i \sim \mathcal{U}(a_i,b_i)$, where $[a_i;b_i]$ are parameter bounds for $p_i$ given by the environment.

\vspace*{1mm}
\noindent  \textbf{Trajectory tracking in simulation.} 
The simulation scene is created based on the sampled parameters. The simulated tool executes the trajectory extracted from the video ten times and the average goal reward is recorded.
We repeat the sampling procedure at most 20,000~times and store the first $K$ parameters $\mathcal{P}_i \sim \mathbb{P}$ that lead to non-zero goal reward~$r^g$. Next, we select $\mathcal{P}_i$ that leads to maximal non-zero goal reward~$r^g$ after executing the robot trajectory.

\subsection{Learning tool manipulation policy in the simulation}
\label{subsec:policy_learning}
The output of the alignment, described in the previous section, are $K$ candidate parameter sets $\mathcal{P}_i$ that adjust the simulated environment to approximately resemble the setup from the input video. The next goal is to use this simulated environment to find a good control policy for the robot. This is a  challenging task because the reward signal provided by the environment is sparse. To overcome this challenge, we will show how to use the aligned tool demonstration as an initialization for the robot policy search.   

We formulate the policy search as a reinforcement learning problem with a fixed time horizon~$H$. The objective function $J(\bm\theta)$ that we wish to maximize is the finite-horizon expected return defined as
\begin{equation}
\begin{aligned}
    J(\bm\theta)  = \mathbb{E}_{\bm\tau \sim \pi_{\bm\theta}} \left[ \sum\nolimits_{t = 0}^H \gamma^t r_t  \right] \, ,
    \label{eq:rl_return}
\end{aligned}
\end{equation}
where $\bm \tau$ is the trajectory generated by the policy $\pi_{\bm\theta}$ parameterized by~$\bm \theta$, $r_t$ is the reward at time $t$ that is composed of the sparse goal reward $r^g_t$ and a penalty for exceeding the joint velocity limits,
and $\gamma$ is a discount factor, set to 0.999 in our experiments, that influences the importance of the reward in time.
The policy~$\pi_{\bm\theta}(\bm{a_t} | \bm{s_t})$ is a neural network that computes action~$\bm a_t$ based on the observed state~$\bm s_t$.
In our setup, action $\bm{a_t} \in \mathbb{R}^N$ is the velocity vector of the robot joints, where $N$ is the total number of degrees of freedom of the robot.
State $\bm{s_t} \in \mathbb{R}^{M}$ consists of the position~($\mathbb{R}^{3}$) and the quaternion representation of orientation~($\mathbb{R}^{4}$) of the tool together with the robot joint positions~($\mathbb{R}^{N}$ for a robot with $N$ degrees of freedom) and the time variable (\ie~$M = 7 + N + 1$).

\noindent  \textbf{Guiding policy search via trajectory optimization.}
Our sparse reward problem is challenging to solve and reinforcement learning is unlikely to find a policy without a good initialization. The objective is to use the tool trajectory extracted from the video and aligned with the simulation as an initialization for the robot policy. This is, however, challenging as the tool trajectory is in Cartesian coordinates and cannot directly be used as a policy initialization as the policy operates in the joint space of the robot. In addition, the position of the robot in the environment is a priori unknown and can greatly affect the difficulty and feasibility of the resulting manipulation problem. 
To address these challenges, we formulate the following trajectory optimization problem to find (i) the sequence of control velocities of the robot $\bm{v_0}^*,\ldots,\bm{v_T}^*$ and (ii) the robot base mounting position $\bm b^*$ that follow the trajectory of the tool extracted from the video:
\begin{equation}
\begin{aligned}
\bm b^*, \bm v_0^* , \dots , \bm v_T^*
  =
    \argmin_{\bm b, \bm v_0, \dots, \bm v_T}{\sum_{t=0}^T{c_t(\bm b, \bm q_t, \bm v_t) }}\\
    \text{s.t. } \bm q_t = \bm q_{t-1} + \bm v_t \Delta t \, ,
    \label{eq:optimization_problem}
\end{aligned}
\end{equation}
where the initial joint position $q_0$ is given and constant, and cost $c_t$ is computed as
\begin{align}
    c_t(\bm b, \bm q_t, \bm v_t) =  
    d(\bm b, \bm q_t, \bm p_t) + 
    w_v \bm v_t^\top \bm v_t +
    w_b c_b(\bm q_t)
     \, , \label{eq:joint_optimization}
\end{align}
where $d(\bm b, \bm q_t, \bm p_t$) denotes the squared distance between the tool pose computed by forward kinematics using the base mounting~$\bm b$ together with the robot configuration~$\bm q_t$ and the target tool pose $\bm p_t$ obtained from the alignment step, the term $\bm v_t^\top \bm v_t$ regularizes the velocity giving preference to smaller velocities,  
function $c_b(\cdot)$ represents a barrier function that penalizes the violation of the joint limits, 
scalar weight~$w_v$ controls the velocity regularization and scalar weight~$w_b$ controls the stiffness of the barrier function.
The constraint $\bm q_t = \bm q_{t-1} + \bm v_t \Delta t$ in~\eqref{eq:optimization_problem} represents system dynamics, where $\bm q_t$ and $\bm q_{t-1}$ are the configurations of the robot at time $t$ and $t-1$, respectively, $\bm v_t$ is the joint velocity of the robot at time $t$ and $\Delta t$ is the time step.
We use Differential Dynamic Programming~(DDP)~\mbox{\cite{crocoddyl20icra}} together with Pinocchio~\cite{carpentier2019pinocchio} to find the fixed robot base position  and the sequence of joint velocities that minimize the cost~\eqref{eq:joint_optimization}.

\begin{figure*}
     \centering
     \vspace*{5pt}
     \begin{subfigure}[b]{0.32\textwidth}
         \centering
         \includegraphics[width=\textwidth]{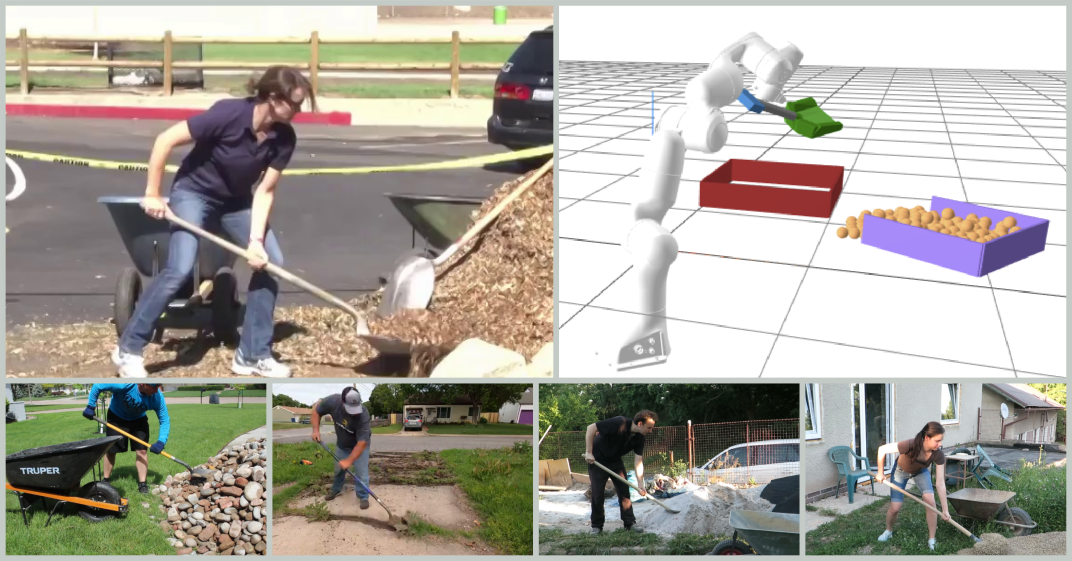}
         \caption{Spade environment and input videos}
         \label{fig:simulation_vs_video_spade}
     \end{subfigure}
     \hfill
     \begin{subfigure}[b]{0.32\textwidth}
         \centering
         \includegraphics[width=\textwidth]{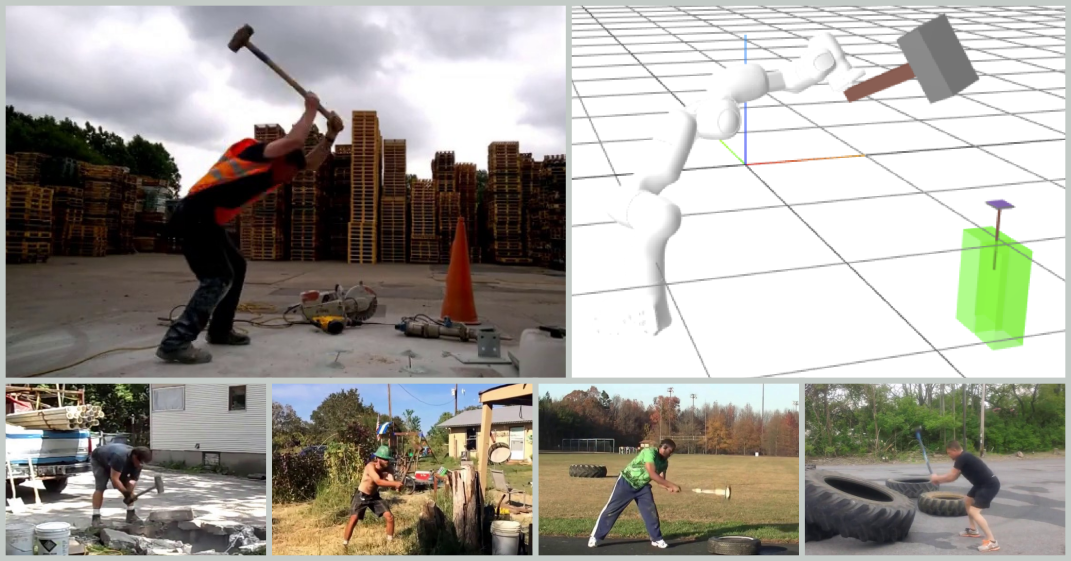}
         \caption{Hammer environment and input videos}
         \label{fig:simulation_vs_video_hammer}
     \end{subfigure}
     \hfill
     \begin{subfigure}[b]{0.32\textwidth}
         \centering
         \includegraphics[width=\textwidth]{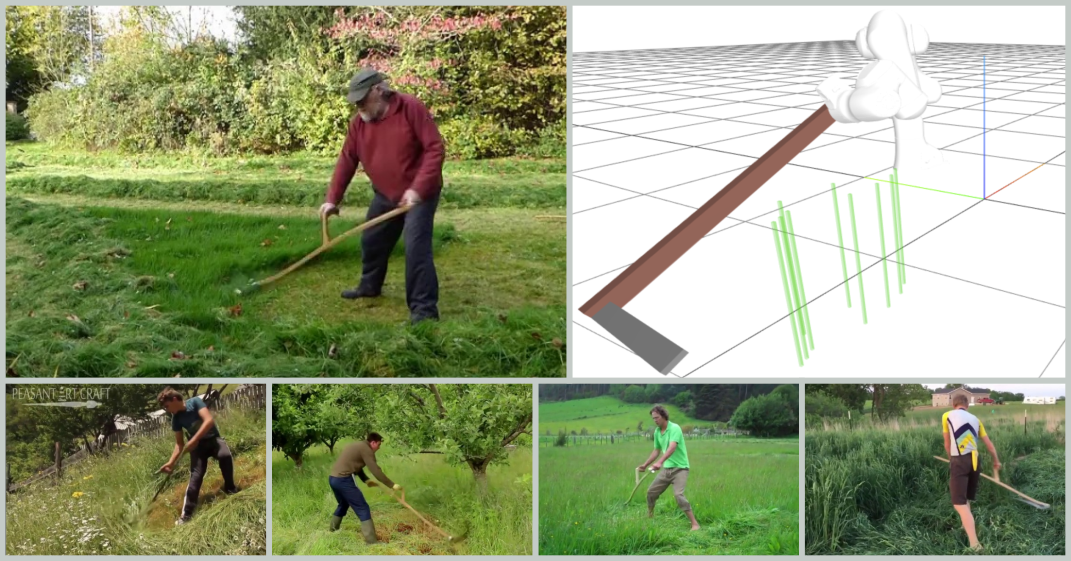}
         \caption{Scythe environment and input videos}
         \label{fig:simulation_vs_video_scythe}
     \end{subfigure}
        \caption{\textbf{Input videos and the corresponding environments for learning the robot manipulation policy for three different tools: (a) spade, (b) hammer, (c) scythe. }
        Each figure shows five different input videos and the input simulation environment that is automatically aligned with each input video using our approach. 
        }
        \label{fig:simulation_vs_video}
\end{figure*}
The described procedure is repeated for each of the candidate parameter set $\mathcal{P}_i$ from the alignment stage. 
The resulting sequence of robot joint velocities, $\bm v_0, \dots, \bm v_T$ is executed in the simulated environment parameterized by $\mathcal{P}_i$. 
We select the set of parameters $\mathcal{P}_i$ that leads to a maximal non-zero goal reward  $r^g$.
The corresponding sequence of robot joint velocities is used to compute state-action pairs ($\bm s_t$, $\bm a_t$) to train an initial policy via behavior cloning~\cite{pomerleau1989alvinn}. 
This initial policy is further fine-tuned by the proximal policy optimization RL algorithm~\cite{ppo} that maximizes the reward defined in~\eqref{eq:rl_return}.  
These steps are crucial for the success of our approach, as will be shown in the experiments (Sec.~\ref{sec:experiments}).

\noindent \textbf{Automatic domain randomization (ADR).} So far, the policy has been learned for a fixed state of the simulated environment found during the alignment step.
To generalize the learned policy for different scene setups, we extend state $\bm s_t$ with the position of the objects and sample these positions during training via automatic domain randomization (ADR)~\cite{openai2019solving}.
This approach gradually extends the scope of the learned policy by gradually increasing the ranges of possible object positions within the environment. The outcome is a policy that generalizes to different positions of  objects in the environment as will be shown in Sec.~\ref{sec:experiments}.

\definecolor{color1}{RGB}{158, 202, 255}
\definecolor{color2}{RGB}{253, 180, 98}
\definecolor{color3}{RGB}{161, 217, 155}
\definecolor{color4}{RGB}{251, 128, 114}
\definecolor{color5}{RGB}{165, 165, 165}

\section{Experiments}
\label{sec:experiments}
\begin{figure*}
    \centering
    \vspace*{7pt}
    \includegraphics[width=\linewidth]{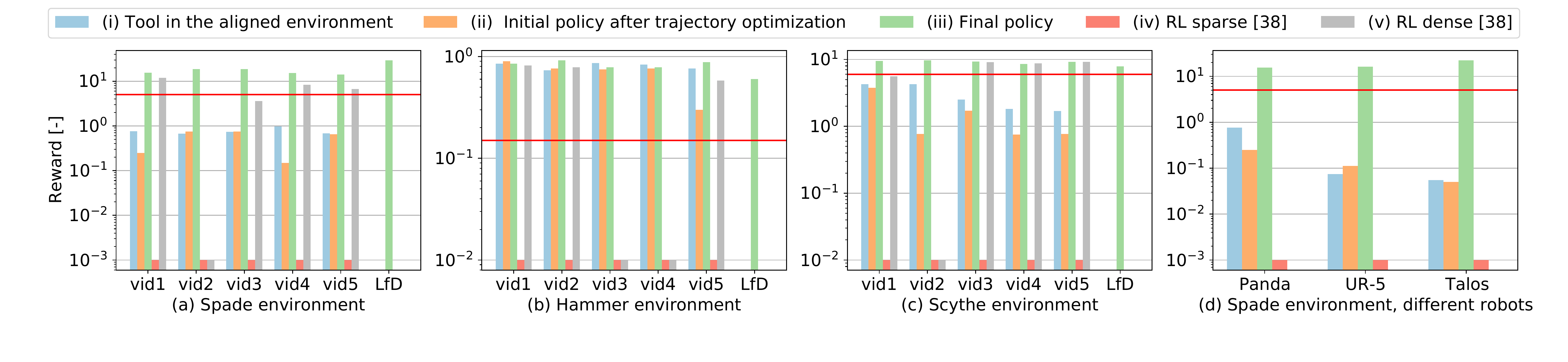}
    \caption{
    \textbf{(Log scale) The goal reward obtained in individual environments.}
    The bar plots show the total goal reward $r^g$ obtained after different stages of our approach:
    (i)~tool motion (without robot) transferred from the video to the aligned environment~\textcolor{color1}{(blue)},
    (ii)~initial robot policy after trajectory optimization~\textcolor{color2}{(orange)},
    (iii)~final robot policy after RL~\textcolor{color3}{(green)}. Results are compared to 
    (iv)~baseline RL approach using only the sparse reward from the environment \textcolor{color4}{(red)} and (v)~baseline RL approach with a manually engineered dense reward~\textcolor{color5}{(gray)}. 
    We learn a separate policy for each input video. Note, that the baseline (iv) did not find a successful policy in any of the experiments and leads to a zero reward, represented by a small bar in the plot. Results are shown for five different videos for the (a)~spade, (b)~hammer, and (c)~scythe environment. 
    For each environment we also report results of learning from kinesthetic demonstration (the last column denoted \textit{LfD}). Plot~(d) shows results for the spade task with three different robots: Panda, UR-5 and Talos. The red horizontal line in each plot shows the success threshold for the given environment.
    \vspace{-5mm}
    }
    \label{fig:main_results_log_scale}
\end{figure*}

We demonstrate the proposed approach on the following three tasks: (i) transferring sand-like material with a spade, (ii) hammering a nail, and (iii) cutting grass with a scythe.
The spade task is also demonstrated for three different robots: (a) Franka Emika Panda, (b) UR5, and (c) the standing Talos robot with a fixed lower body. The learned spade manipulation policy for Franka Emika Panda is demonstrated on the real robot.
 In the following, we provide the details of the simulated environments (Sec.~\ref{subsec:simulated_envs}), outline the structure of the learned policies (Sec.~\ref{subsec:policy_structure}), and provide the quantitative and qualitative evaluation of the learned policies (Sec.~\ref{subsec:evaluation}). {\bf Please see additional results including real robot experiments in the supplementary video available at~\cite{project_page}.}  

\subsection{Simulated environments}
\label{subsec:simulated_envs}
All physics simulations are performed in the PhysX engine~\cite{physx}, with the control frequency set to 24~Hz.
We have created three simulated environments: for the spade, hammer and scythe tools.
In each environment we include a negative reward for joint velocities above the limits of the robot.

\noindent\textbf{Spade environment.}
The goal in the spade environment is to transfer sand from the sand deposit to the desired position called the goal box. The input videos depicts a human holding the spade and transferring leaves or gravel from a pile to a burrow, as shown in Fig.~\ref{fig:simulation_vs_video_spade}. The simulated environment for this task includes the spade tool, the goal box, and the deposit with sand-like material.
We approximate sand with a collection of spheres, and the sand deposit is formed by three walls, which prevent free movement of the spheres.
The placement of the scene elements and sand box orientation are found automatically by the proposed alignment procedure, described in Sec.~\ref{subsec:trajectory_alignment}, such that the environment approximately resembles the input video.
The control policy is trained in the aligned simulated scene and is then executed on the real robot.
The environment for the real robot experiment is built to match the simulated scene.
The goal reward $r_t^g$ in the spade environment is defined as the number of spheres delivered to the goal box. 

\noindent\textbf{Hammer environment.} The goal in the hammer environment is to plant the nail. The five input videos depict a human using a hammer to break the concrete or to hit a tire, as shown in Fig.~\ref{fig:simulation_vs_video_hammer}. The simulated environment for this task includes the hammer tool and the target nail object. The nail object is connected to the ground plane and constrained to move only in the $z$ direction and only in the range of 0 to 100~mm. 
The goal reward $r_t^g $ for the hammer environment is computed based on the position of the nail along $z$-axis denoted as~$d_t^z$.
The reward is one, if the target is nailed completely ($d_t^z < 0.001$) and zero otherwise.
The reward encodes whether or not the nail has completely entered into the material. 

\noindent\textbf{Scythe environment.} The goal in the scythe environment is to cut grass. The five input videos depict a human using a scythe to cut grass. Sample frames from the input videos are shown in Fig.~\ref{fig:simulation_vs_video_scythe}. The simulated environment for this task includes the scythe tool and patch of several grass elements, where each grass element is represented by a thin vertical cuboid. Grass is randomly generated inside the patch. The placement of the grass patch is found automatically by the proposed alignment procedure 
(Sec.~\ref{subsec:trajectory_alignment})
to approximately resemble the input video scene.
The goal reward $r_t^g $ for the scythe environment is defined as the number of cut grass elements. 
A grass element is considered as cut if it is intersected by the tool blade close to the ground plane (\ie the $z$-coordinate of the intersection point is less than a predefined threshold~$z_{max}$) and the speed of the blade in the cut direction is higher than a pre-defined threshold~$v_{min}$.

\subsection{Policy structure}
\label{subsec:policy_structure}
The policies for the robot were trained with the Proximal Policy Optimization algorithm~\cite{ppo}. For the policy, we use a neural network with two fully-connected hidden layers that consist of 400 and 300 neurons, respectively, and have ReLU activations. The input to the policy is state vector $\bm s_t$. The output of the policy is the mean value $\mu(\bm s_t)$ of the action distribution. The action $\bm{a_t} \in \mathbb{R}^N$ is the velocity vector for the robot joints, where $N$ is the number of degrees of freedom.  The action is sampled from the Gaussian distribution $\bm a_t~\sim~\mathcal{N}(\mu(\bm s_t), \Sigma)$, where the covariance matrix $\Sigma$ is diagonal with learnable elements.
The value function approximator is a neural network with two hidden layers consisting of 400 and 300 neurons with ReLU activations.

\subsection{Evaluation}
\label{subsec:evaluation}
\noindent\textbf{Quantitative evaluation.}
We compare the performance of the following policies:
\noindent\textit{(i)~tool in the aligned environment} that imitates the tool motion extracted from the video in the aligned scene~(Sec.~\ref{subsec:trajectory_alignment} - tool motion without a robot),
\textit{(ii)~initial robot policy after trajectory optimization} as described in Sec.~\ref{subsec:policy_learning},
\textit{(iii)~final robot policy} learned with the proposed approach, \textit{(iv)~baseline sparse RL approach} that learns a robot policy using PPO~\cite{ppo} with only the sparse goal reward given by the environment without using the video to initialize the learning , and \textit{(v)~baseline RL approach with manually engineered dense reward}.
For the spade environment, dense reward consists of the exponential distance between (a) the tool and the sand deposit and (b) the exponential distance between the pile of sand and the goal box. 
For the hammer environment, the dense reward consists of the exponential distance between the tool and the nail. For the scythe environment, we define two 3D points $\bm x^A$ and $\bm x^B$ that are located on the opposite sides of the grass patch at the ground level. The dense reward includes a positive reward for the tip of the tool being close to the point $\bm x^A$  in the first half of the trajectory and to point $\bm x^B$  in the second half of the trajectory. Also, we add a reward for keeping the tool rotation close to the reference rotation (scythe parallel to the ground). Additional details about computing the dense reward are in the extended version of this work available at~\cite{project_page}.
Moreover, for each tool type we conduct an experiment where we learn a policy from an expert kinesthetic demonstration. We record a trajectory of an expert manipulating the robot to complete the task and use this trajectory to train a policy via behavior cloning~\cite{pomerleau1989alvinn}. This policy is further finetuned by PPO~\cite{ppo} and the resulting reward is reported in Fig.~\ref{fig:main_results_log_scale} as \textit{LfD}. The performance of the policies learned from the expert kinesthetic demonstrations are comparable to results obtained with our method which confirms our hypothesis that we can use the video instead of a kinesthetic demonstration to learn a tool manipulation policy. Please refer to the supplementary video for the visualization of the obtained trajectories.
The quantitative evaluation of the policies is shown in Fig.~\ref{fig:main_results_log_scale}.
Along with reporting goal reward $r^g$ obtained by the final policy, we also define the following notion of success: (a) spade environment - at least 10 spheres in the goal box for at least half of the episode, (b) hammer environment - the nail is planted in the first second, (c) scythe environment - all grass elements in the patch are cut in the first second.  With this definition of a success rate, the proposed approach achieves 100\% success for all environments. 

Note that different videos result in a different alignment with the simulated environment and therefore have different intermediate, (i)+(ii), and final, (iii), results.
Ideally, the reward gained after repeating the tool motion without the robot in the aligned environment (i) and executing the initial policy with the robot (ii) should be the same. The small differences are caused by the robot kinematics constraints that do not allow the initial robot policy (ii) to exactly follow the aligned tool trajectory (i) from the video. 
Also, we penalize velocity limit violation in the RL step and the resulting policies respect the joint velocity limits of the robot, but might take more time to solve the task. This is reflected by the performance drop in the hammer environment where step~(i) of the approach has a higher reward than the final policy, which respects the velocity limits.

For our approach, it is crucial that repeating the tool motion in the aligned environment (i) gets a non-zero reward.
A non-zero reward is preserved by the initial robot policy obtained by trajectory optimization (ii) as the optimization procedure is designed to follow the aligned tool trajectory.
The final policy (iii) learned by RL obtains the highest reward in most of the cases. 
Frames from the trajectory produced by the spade final policy are shown in~\cref{fig:teaser} and other final policies are depicted in the companion video.
The superior performance of the final policy (iii) could be attributed to the meaningful policy initialization as a result of alignment (i) and trajectory optimization (ii) steps, which can effectively guide the learning process to solve the underlying sparse reward task.
Without good initialization, the agent does not obtain any reward and performs a random blind search in the environment.
This random search is unlikely to find a proper policy for tasks with sequential dependencies, for example, in the spade environment the agent needs to grab the spheres before transferring them into the goal box.
This is confirmed by our sparse reward baseline~(iv), which obtains zero reward for all environments.
The dense reward baseline~(v) achieves performance that is comparable to the proposed approach, however, only for 60\% of the alignments for each task.
In addition, the dense reward needs to be manually engineered and fine-tuned for each task separately.

\begin{figure}
    \centering
    \vspace*{7pt}
    \includegraphics[width=\linewidth]{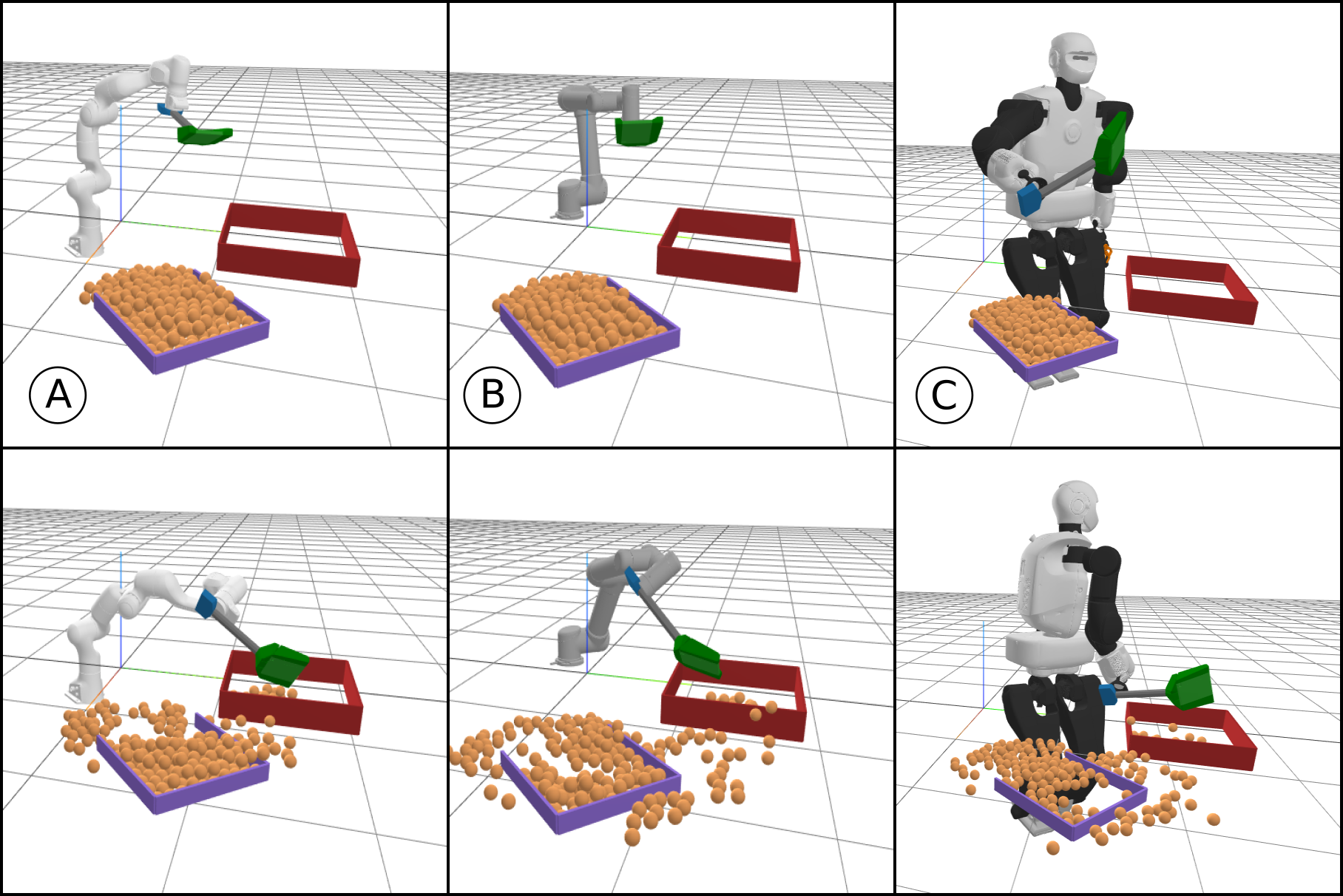}
    \caption{
    \textbf{Visualization of the learned policies for different robot morphologies.}
    Three different robots were used in our experiments: (A)~Franka Emika Panda, (B)~UR5 robot, and (C)~standing Talos robot with a fixed lower body. Frames are chosen to represent similar stage of motion that may occur at different timesteps for different policies.
    For Franka Emika Panda robot, sand and goal box are elevated to correspond to the real experiment (Fig.~\ref{fig:teaser}). For the Talos robot, sand and goal box are elevated to knee level.
    }
    \label{fig:diff_robots_visual}
\end{figure}

\noindent\textbf{Adaptation to different robot kinematic structures.}
Using the tool trajectory instead of the human motion allows us to be agnostic to the robot morphology when we transfer between the input video and simulation. 
This allows us to easily train new policies for other robots with different morphologies without the need for motion re-targeting, which is one of the key limitations of the current state-of-the-art methods~\cite{DBLP:journals/corr/abs-1804-02717, RoboImitationPeng20, 2018-TOG-SFV}.
To demonstrate this capability, we have trained a new policy for the spade task using two additional robots: the 6~DoF UR-5 manipulator and standing Talos robot with a fixed lower body, which has 11~DoF: 2 for the torso and 9 for the right arm.
Note that Franka Emika Panda robot that is used in all other experiments has 7~DoF. 
The quantitative evaluation of the spade task policies for the three mentioned robots is shown in Fig.~\ref{fig:main_results_log_scale}~(d).
The results show that all robots achieved a similarly high reward after learning via the proposed approach.
The visualization of the learned policies is shown in Fig.~\ref{fig:diff_robots_visual}.

\noindent\textbf{Policy generalization.}
We have applied  ADR (Sec.~\ref{subsec:policy_learning}) in the spade environment and learned a policy for the goal box position randomly sampled from a 100x90~cm rectangle around the initial position.
The performance of the policy for different box placements is shown in Fig.~\ref{fig:heatmap_spade_ADR} and in the supplementary video. 

\begin{figure}[!b]
  \centering
  \vspace*{-7pt}
  \includegraphics[width=\linewidth]{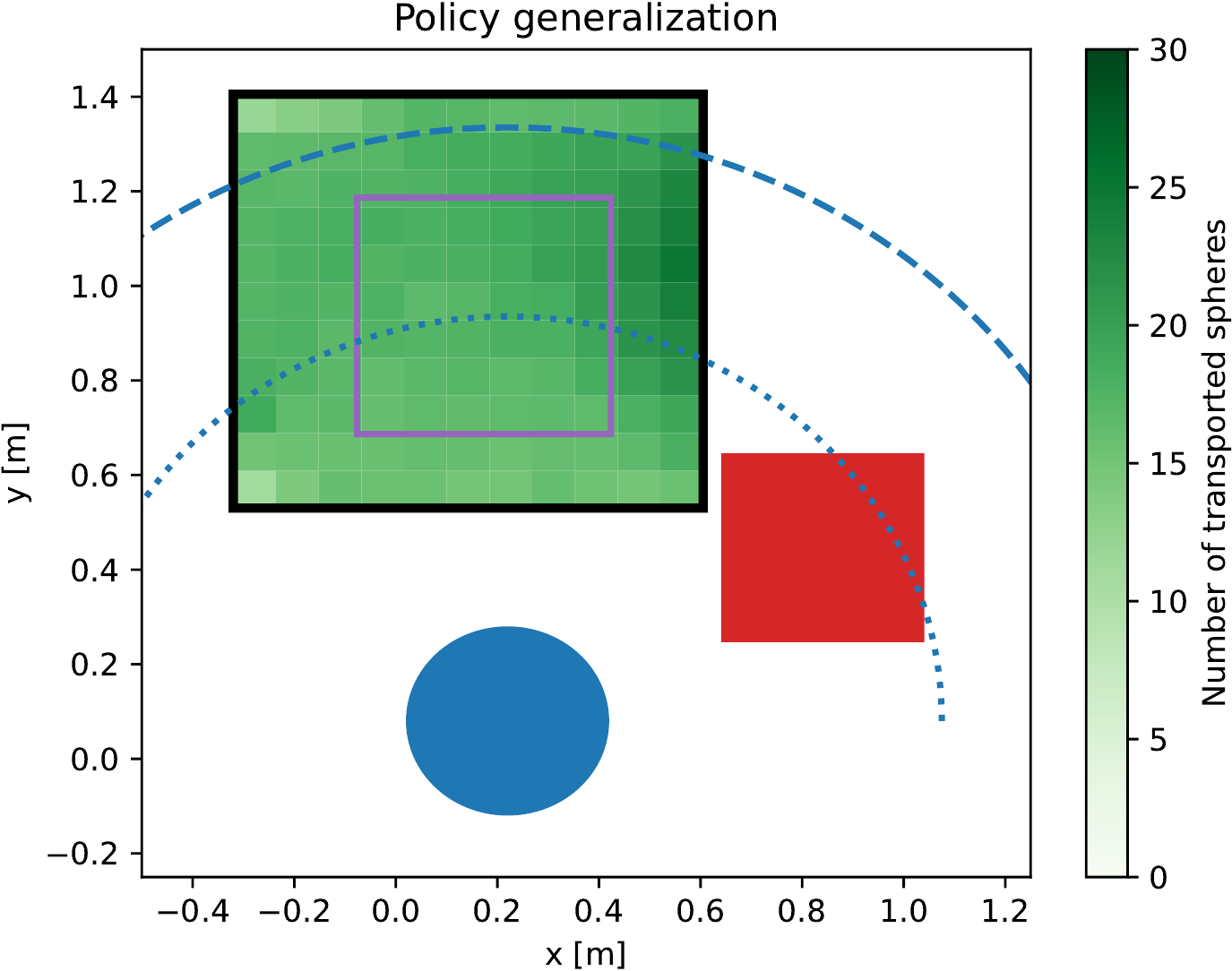}
  \caption{
  \textbf{Policy generalization in the spade environment.} 
  We evaluate the policy learned by the automatic domain randomization (ADR) for various goal box positions (green cells correspond to the different centers of the goal box).
  The colorbar on the right shows the number of spheres transferred to the goal box.
  The initial position of the goal box (purple) is used as a starting point for ADR, which generalizes the policy to cover the range of goal box positions specified in black.
  The sand deposit (red) and the robot base (blue) are fixed.
  The blue curves show the robot working area (dotted) and the working area extended by the tool length (dashed).
  }
  \vspace*{-5pt}
  \label{fig:heatmap_spade_ADR}
\end{figure} 

\noindent\textbf{Real robot experiments.} 
For the real robot experiments, we use the 7~DoF Franka Emika Panda robotic arm~\cite{franka-panda-robot}.
As discussed in Sec.~\ref{subsec:policy_learning}, inputs and outputs of the policy represent only kinematic quantities. Therefore, we can compute the joint trajectory of the robot offline by using the forward kinematics of the robot.
This trajectory is then executed on the real robot in an open-loop manner.
According to the success rate defined for the simulated experiments, we achieve 100\% success rate for the real-world experiment.
The visualization of the robot controlled by the policy is shown in Fig.~\ref{fig:teaser}.
Please see the supplementary video for the recording.

\noindent\textbf{Limitations and assumptions.} 
Our approach has several limitations. 
First, we assume that the simulated environment with a sparse reward is known and contains only task-related objects whose positions are found during the alignment stage. 
Second, we assume that the tool type is known (spade, hammer, or scythe) and the selected input videos feature a human manipulating a tool of that type. The tool motion reconstruction module [7] uses this tool type information for training the tool detection module and a simplified stick-like 3D model of the tool for the tool 3D motion estimation. 
Third, we assume the human and the manipulated tool are well visible in the video without significant (self-)occlusions. 
Fourth, we trim only one cycle of motion for tasks with sequential dependency (\ie spade).

\section{Conclusion}
\label{sec:conclusion}
We have presented an approach for learning robot-tool manipulation by watching instructional videos. From the video, we extract the 3D tool trajectory and propose an alignment procedure to find a suitable simulation state that approximates the real-world setup observed in the video.
We use the alignment procedure followed by trajectory optimization to initialize the control policy search in the simulated environment.
Our evaluation shows that learning such policies is nontrivial, and the video demonstration is essential for success. 
By leveraging the video, we overcome the need for costly manual demonstration in the robot's environment.
This opens up the possibility of considering a wider range of target tools, tasks, and robots including moving platforms and humanoids.

\bibliographystyle{IEEEtran}
\bibliography{IEEEabrv,example}
\appendix

\section{Dense reward details}
\label{sec:dense_reward}
\noindent\textbf{Manually designed dense reward.}
In this appendix we provide details of computing the dense reward that we manually designed for the baseline RL approach (Sec.~\ref{subsec:evaluation}).
For the spade environment, the dense reward consists of the exponential distance between 
(a) the tool tip position $\bm x^T$ and the sand deposit $\bm x^S$, 
and (b) the exponential distance between the pile of sand ($K$ spheres with positions $\bm x^1 \dots \bm x^K$) and the goal box position $\bm x^G$: 
\begin{equation}
\begin{aligned}
    r_t^{d}~=~\frac{1}{H}\exp\left(-\frac{b}{2} \, d( \bm x^T_t, \bm x^S_t) \right)+ \\
    \sum_{i = 1} ^ K{\frac{1}{H K}\exp\left(-\frac{b}{2} \, d( \bm x^G_t, \bm x^i_t) \right)} ,
    \label{eq:dense_spade}
\end{aligned}
\end{equation}
where
$H$ is the environment horizon,
$b$ is the reward length-scale parameter
and function~$d(\cdot, \cdot)$ measures the Euclidean distance between two 3D points.

For the hammer environment, the dense reward consists of the exponential distance between the tool tip position $\bm x^T$ and the nail position $\bm x^N$: 
\begin{equation}
\begin{aligned}
    r_t^{d}~=~\frac{1}{H}\exp\left(-\frac{b}{2} \, d( \bm x^T_t, \bm x^N_t) \right),
    \label{eq:dense_spade}
\end{aligned}
\end{equation}
where
$H$ is the environment horizon,
$b$ is the reward length-scale parameter
and function~$d(\cdot, \cdot)$ measures the Euclidean distance between two 3D points.

For the scythe environment, we define two 3D points $\bm x^A$ and $\bm x^B$ that are located on the opposite sides of the grass patch at the ground plane. 
The dense reward includes 
(a) exponential distance between the tip of the tool $\bm{x}^T$ and the point $\bm x^A$  in the first half of the trajectory, 
(b) exponential distance between the tip of the tool $\bm{x}^T$ and the point $\bm x^B$  in the second half of the trajectory, and
(c) exponential distance between the tool rotation $\bm q^T$ and reference rotation $\bm q^R$ (scythe is parallel to the ground):
\begin{equation}
\begin{aligned}
    r_t^{d}~=~\mathds{1}_{t<H/2}\frac{1}{H}\exp\left(-\frac{b}{2} \, d( \bm{x}^T_t, \bm{x}^A) \right) +\\ 
    \mathds{1}_{t\geq H/2}\frac{1}{H}\exp\left(-\frac{b}{2} \,  d( \bm{x}^T_t, \bm{x}^B) \right) + \\
    \frac{1}{H}\exp\left(-\frac{b}{2} \, d^I( \bm q^T_t, \bm q^R_t) \right),
    \label{eq:dense_spade}
\end{aligned}
\end{equation}
where
$H$ is the environment horizon,
$b$ is the reward length-scale parameter,
function~$d(\cdot, \cdot)$ measures the Euclidean distance between two 3D points,
and function~$d^I(\bm q_1, \bm q_2)$ measures the intrinsic distance between two rotations which is computed as a positive length of the geodesic arc connecting $\bm q_1$ to $\bm q_2$.

\end{document}